\pdfoutput=1

\documentclass[11pt]{article}

\usepackage{ACL2023}

\usepackage{times}
\usepackage{latexsym}

\usepackage{graphicx}
\usepackage[normalem]{ulem}
\useunder{\uline}{\ul}{}

\usepackage[T1]{fontenc}

\usepackage[utf8]{inputenc}

\usepackage{microtype}

\usepackage{inconsolata}
\usepackage{scalefnt}

\title{Conversational Topic Recommendation in Counseling and Psychotherapy with Decision Transformer and Large Language Models}

\author{Aylin Gunal \\ \scalefont{0.9} University of Michigan \\ \scalefont{0.9} Ann Arbor, MI \\ \texttt{gunala@umich.edu}
        \And
        Baihan Lin \\ \scalefont{0.9} Icahn School of Medicine at Mount Sinai \\ \scalefont{0.9} New York, NY \\ \texttt{baihan.lin@mssm.edu} \And
  Djallel Bouneffouf \\ \scalefont{0.9} IBM Research \\ \scalefont{0.9} Yorktown Heights, NY\\ 
  \texttt{djallel.bouneffouf@ibm.com} \\}

\begin{document}
\maketitle
\begin{abstract}
Given the increasing demand for mental health assistance, artificial intelligence (AI), particularly large language models (LLMs), may be valuable for integration into automated clinical support systems. In this work, we leverage a decision transformer architecture for topic recommendation in counseling conversations between patients and mental health professionals. The architecture is utilized for  offline reinforcement learning, and we extract states (dialogue turn embeddings), actions (conversation topics), and rewards (scores measuring the alignment between patient and therapist) from previous turns within a conversation to train a decision transformer model. We demonstrate an improvement over baseline reinforcement learning methods, and propose a novel system of utilizing our model's output as synthetic labels for fine-tuning a large language model for the same task. Although our implementation based on LLaMA-2 7B has mixed results, future work can undoubtedly build on the design.
\end{abstract}

\section{Introduction}

In recent years, there has been a notable uptick in the number of people seeking professional help for mental health concerns, but the available pool of mental health professionals remains small in comparison. To address this need, automated AI-based tools and methods for counseling have been explored and engineered, ranging from systems for training junior mental health counselors \cite{Min2022PAIRPM, Demasi2019TowardsAC} to AI-in-the-loop chatbots \cite{Sharma2022HumanAICE}.
With the dramatic rise in popularity and accessibility of large language models (LLMs), it's expected that LLMs will play a significant role in the intersection of computing and mental health research, as well. 

In our prior work \cite{lin2023supervisorbot}, we introduced the SupervisorBot, a reinforcement learning (RL)-based topic recommendation system in counseling conversations. 
This proves to be a useful tool for clinicians during their psychotherapy sessions, where the system recommends what topics to discuss next given what has been discussed so far, as well as what works best in the past in terms of the patient outcomes.
In this work, we improve upon this meaningful task by introducing the Decision Transformer \cite{Chen2021DecisionTR}, a transformer model designed for reinforcement learning (RL), into the recommendation pipeline, demonstrating better performance than other RL methods. We also explore the potential combination of Decision Transformer with LLMs, by generating labels for unseen transcript data using the pre-trained Decision Transformer model, and feeding the synthetically annotated data to fine-tuning a LLM. 
Our primary contribution is demonstrating that in the task of topic recommendation, Decision Transformer outperforms baseline RL methods; if such a system were to go through the process of user testing, the Decision Transformer––or models building on or improving Decision Transformer––can be utilized as the backbone for the recommendation module.

We first describe how we implement the pre-processing of the therapy conversation dataset, and how this is fed into the Decision Transformer model. We then describe how we use a portion of the dataset to train the Decision Transformer model, and that trained model's predicted labels are used as input to a large language model to train for the same task of topic recommendation.

\section{Related Work}

Decision Transformer was introduced as a transformer-based architecture to abstract the process of offline reinforcement learning, and has been used successfully in various NLP tasks including natural language understanding \cite{zhang2022can, bucker2023latte}, navigating text-based games \cite{putterman2021pretraining}, and generative language modeling \cite{memisevic2022decision}. The Decision Transformer architecture has also been effectively applied to the clinical domain to generate treatment recommendations based on patient history \cite{lee2023clinical}. 
In this work, we effectively apply the Decision Transformer architecture to the mental health domain in a dialogue recommendation task and improve on performances with older reinforcement learning methods.

The improvement of AI-in-the-loop tools to support humans in tasks has typically focused on human feedback, although more recent work has explored the potential for AI tools to improve themselves through a number of methods. \cite{Saunders2022SelfcritiquingMF} demonstrates that a generative language model can improve its own outputs through fine-tuning on its own generations, and that the improvements are more significant as the model size increases. Generative models also have the advantage of being able to generate improvements to their own outputs \cite{Zelikman2023SelfTaughtO}.

In addition to models improving themselves, ensemble methods in which one model serves some intermediary purpose within the pipeline––e.g. data generation or filtering for input to another model––can be used for conversational modeling tasks as well \cite{huang2023ensemble}. \cite{stiennon2020learning} uses an intermediary model's output as a reward function for another model, outperforming sole supervised learning from the source dataset. In this work, we explore a potential pipeline in which one model's output is used as synthetic data to train a language model for the task of topic recommendation. We consider the idea of AI supplementing a typical reinforcement learning with human feedback (RLHF) process by experimenting with how AI may be able to augment feedback, which can have  significant implications given the lack of publicly available mental health dialogue data, let alone annotated data.

\section{Architecture}

\begin{figure*}[h]
    \centering
    \includegraphics[width=\textwidth]{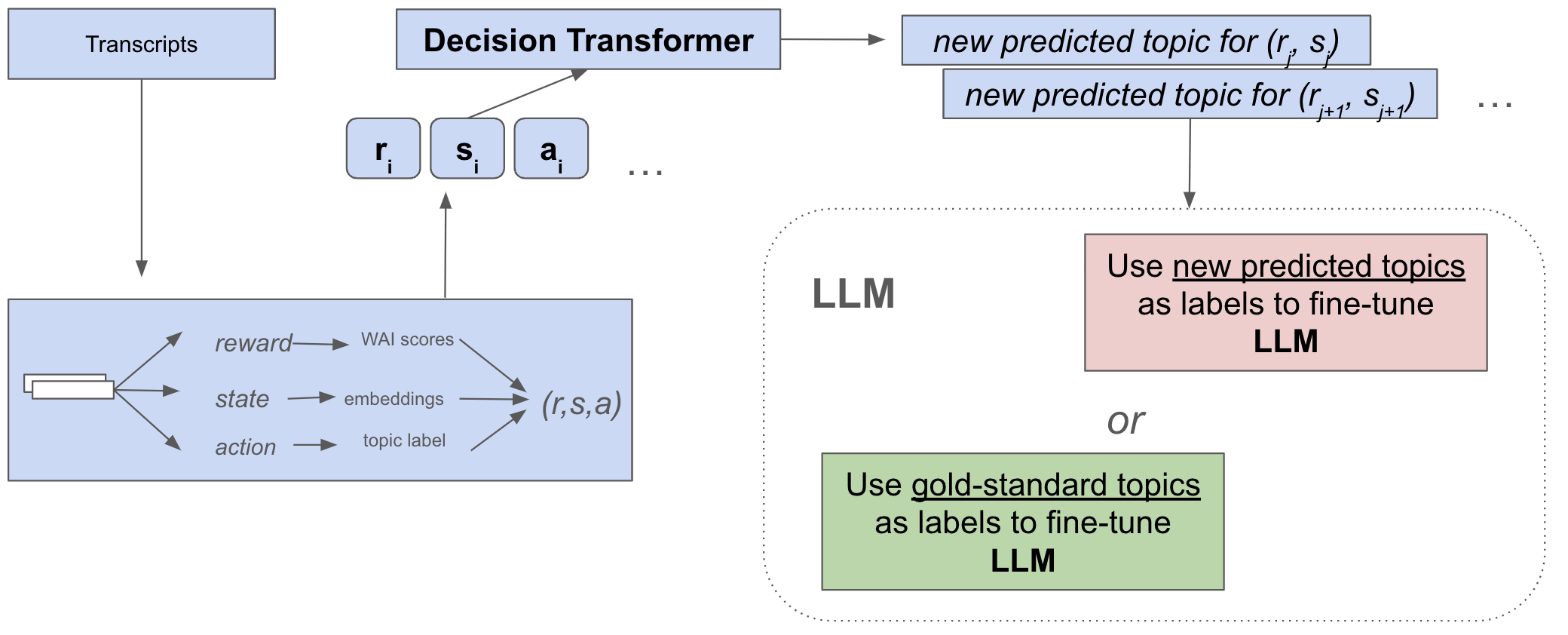}
    \caption{Architecture of the proposed LLM integration, demonstrating how both gold-standard labels from the dataset as well as synthetic annotations from Decision Transformer output can be used to fine-tune the LLM.}
    \label{fig:fig1}
\end{figure*}

In the following sections, we describe the architecture of our system in detail (see Fig. \ref{fig:fig1}).

\subsection{Decision Transformer} \label{sec-dt-experiments}

We re-implement the recommendation system pipeline as described in our original paper, \cite{lin2023supervisorbot}. 
This  system is designed to provide real-time feedback in the form of next-topic recommendation for mental health counselors in session with patients, using reinforcement learning methods to learn and to recommend the next topic (the \textit{action} taken by the counselor) to move on from the current segment of dialogue (the current \textit{state}). \textit{Rewards} are calculated using working alliance inventory (WAI) \cite{Horvath1989DevelopmentAV}, a score from a survey of questions to determine how aligned a counselor is with their patient within a session. WAI is determined by computing similarity between inventory items and segments of dialogue \cite{Lin2022DeepAO}, and inventory items fall under three different categories: Task, Bond, and Goal. We include an aggregate WAI score, referred to as Full.

The original SupervisorBot paper evaluates the system's performance on three baseline RL algorithms: DDPG \cite{lillicrap2015continuous}, TD3 \cite{fujimoto2018addressing}, and BCQ \cite{fujimoto2019off}.

We use the Alex Street dataset \footnote{https://alexanderstreet.com/products/counseling-and-psychotherapy-transcripts-series}, a dataset composed of counseling session transcripts for patients suffering from depression, anxiety, suicidal thoughts, and schizophrenia. 
The Alex Street dataset is preprocessed and segmented into turn-pairs, which are then embedded using Word2Vec \cite{Mikolov2013DistributedRO}. We use embedded topic modeling \cite{dieng2020topic} to extract 8 topics from the corpus––as determined optimal by the motivating paper––and label each turn-pair with the topic it best represents. The WAI scores are computed for each turn-pair. As the original system design is done, turn-pair embeddings represent states, topic labels represent actions, and associated WAI scores represent rewards. These items are fed as input into the Decision Transformer in the form of tuples of $(r_t, s_t, a_t)$.   

We defer to the original Decision Transformer paper for architecture details. Our model contains a single-head, 3-layer attention mechanism, and we use a context window of 20 for the baseline results.
Pearson's correlation between the model predicted actions and real actions taken is used for evaluation for all experiments. We run experiments 5 times using a 95\%/5\% train-test split, and take the average result.

\subsection{LLMs for Recommendation}

An attractive property of LLMs is flexibility in usage; at their core they simply model language probability distributions, making their outputs malleable to various tasks in NLP. In this section, we explore LLMs' ability for dialogue classification into labels of different psychiatric conditions to demonstrate their usefulness in various components of RLHF, similar to a diagnostic scenario in clinical setting as in \cite{lin2022working,lin2024compass}. We primarily experiment with LLaMA-2 with 7B parameters \cite{touvron2023llama}. We fine-tune LLaMA-2 model for sequence classification and test using the same preprocessing steps and train-test split described in Section \ref{sec-dt-experiments}. During the fine-tuning process, we do not use a validation dataset to avoid data leakage into the test set.

We experiment with treating Decision Transformer predictions as synthetic gold standard annotations for the LLM to learn from. We split the full dataset in a 40\%/40\%/20\% split; the first 40\% of the Alex Street dataset is used to train Decision Transformer, then Decision Transformer outputs predictions for another 40\% of the dataset which the LLM is then fine-tuned with, and ultimately the LLM is evaluated on the final 20\% of the dataset. Due to computational constraints, we apply low-rank adaptation (LoRA) \cite{hu2021lora}, a parameter efficient fine-tuning method, in order to optimize the fine-tuning process. The LoRA configuration includes an alpha of 16 and dropout rate of .05, and we fine-tune for 1 epoch. 
We target the LLaMA-2 model's attention layers during training and save the final layer's weights to avoid those scores being randomly initialized for inference. 
To provide some baseline for comparison, we additionally fine-tune the LLaMA-2 model on the original gold-standard labels.

\section{Evaluation}

\begin{table*}[]
\resizebox{\textwidth}{!}{%
\begin{tabular}{lccccc}
{\ul \textbf{Decision Transformer}} &               &               &               &               &               \\
                                    & Depression    & Anxiety       & Schizophrenia & Suicidal      & All           \\ \hline
\multicolumn{1}{l|}{Full}           & .176          & .233          & .246          & .213          & .361          \\
\multicolumn{1}{l|}{Task}           & \textbf{.291} & \textbf{.320} & .247          & .231          & .323          \\
\multicolumn{1}{l|}{Bond}           & .270          & .314          & .231          & \textbf{.239} & .335          \\
\multicolumn{1}{l|}{Goal}           & \textbf{.291} & .313          & \textbf{.249} & .229          & \textbf{.375}
\end{tabular}%
}
\caption{Results of Decision Transformer on topic recommendation task, using previous 20 turn-pairs as input. Best results per data subset are in bold.} 
\label{tab:all-models-results}
\end{table*}

\begin{table}[]
\resizebox{\columnwidth}{!}{%
\begin{tabular}{lcccc}
                           & \multicolumn{4}{c}{\textbf{Context Lengths}} \\ \cline{2-5} 
\textbf{Rewards}           & 5         & 10        & 15        & 20       \\ \hline
\multicolumn{1}{l|}{Full}  & 0.346     & 0.345     & \textbf{0.403}    & 0.361    \\
\multicolumn{1}{l|}{Bond}  & 0.284     & 0.343     & \textbf{0.359}     & 0.335    \\
\multicolumn{1}{l|}{Task}  & 0.272     & 0.298     & \textbf{0.342}     & 0.322    \\
\multicolumn{1}{l|}{Goal} & 0.278     & 0.339     & 0.348     & \textbf{0.375}   
\end{tabular}%
}
\caption{Decision Transformer model performance trained on varying context lengths. Best results per reward scale are in bold.}
\label{tab:ablation-context-lengths}
\end{table}

\subsection{Results}

\begin{table}[]
\centering
\resizebox{\columnwidth}{!}{%
\begin{tabular}{llll}
                          & \textbf{DDPG}        & \textbf{BCQ}          & \textbf{TD3}          \\ \cline{2-4} 
\multicolumn{1}{l|}{Full} & .264 (-.97) & .170 (-1.91) & .286 (-.75)
\end{tabular}%
}
\caption{Baseline RL performance on full-scale rewards on the full dataset, with a comparison to DT performance (the negative values suggest that these baselines perform worse than our proposed DT-based recommendation model, which improves upon them).}
\label{tab:baseline-rl-fulls}
\end{table}

The results of the Decision Transformer on a 95\%/5\% train-test split, reflecting the set-up of the original SupervisorBot paper, are provided in Table \ref{tab:all-models-results}. We reproduce results for the other RL methods in the original paper for performance on the full-scale rewards; Decision Transformer outperforms these baselines as noted in Table \ref{tab:baseline-rl-fulls}. 
We note that Decision Transformer specifically performs best for all reward scales when trained on the full dataset; among individual diseases, the model performs best on the task, bond, and goal scales for anxiety. 

We additionally evaluate whether or not the 20-timestep context is necessary for good performance from the Decision Transformer model, and these results are provided in Table \ref{tab:ablation-context-lengths}. We note that 15 time-steps is optimal for a majority of the reward sclaes, suggesting that the Decision Transformer is better able to make decisions provided a briefer learning history. An advantage of utilizing a transformer-based model for this task is that we are able to investigate its internal structure to understand specifically which historical features––including which time-steps––are significant for inference.

Additionally, we note that the LLaMA-2 model trained on the gold-standard data does not necessarily outperform the Decision Transformer for all reward scales as indicated in Table \ref{tab:ft-llama7b}, indicating that the off-the-shelf language model may not be conducive for a reinforcement learning task. 
LLaMA-2 trained on the Decision Transformer output directly also does not perform particularly well; future work may include modifying the way in which the Decision Transformer synthetic labels are used by a language model. It's possible that prompting the language model may yield better results than treating it as a sequence classifier.

\begin{table}[]
\centering
\resizebox{\columnwidth}{!}{%
\begin{tabular}{lllll}
                                               & Full & Task & Bond & Goal \\ \cline{2-5}
                                               
\multicolumn{1}{l|}{\textbf{LLaMA-2 7B  + DT}} & .148 & .118 & .158 & .115 \\
\multicolumn{1}{l|}{\textbf{LLaMA-2 7B + Gold}} & .371 & .259 & .315 & .332
\end{tabular}
}
\caption{Results of fine-tuning LLaMA-2 7B on DT output and gold-standard labels.}

\label{tab:ft-llama7b}
\end{table}

\subsection{Additional Analysis for Interpretability} \label{sec:additional-analysis}

\begin{figure} 
    \centering
    \includegraphics[width=.95\linewidth]{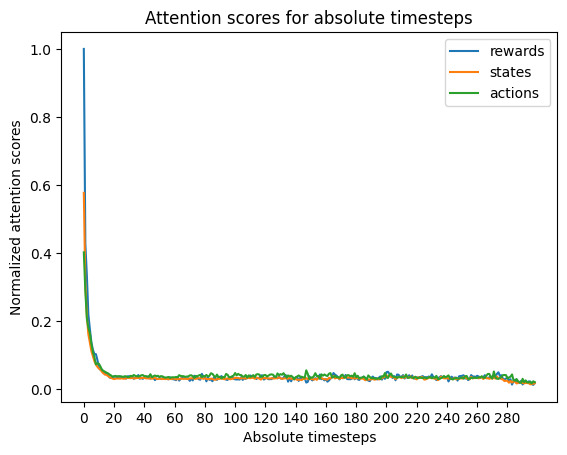}
    \caption{Normalized attention scores associated with absolute timesteps, \textit{without} padded sequences.}
    \label{absolute-timesteps}
\end{figure}

\begin{figure} 
    \centering
    \includegraphics[width=.95\linewidth]{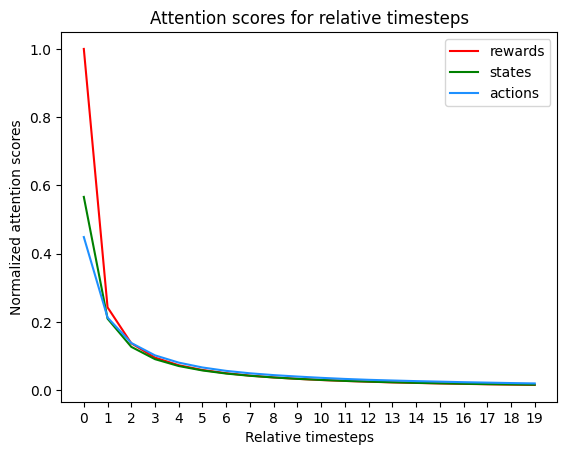}
    \caption{Normalized attention scores associated with relative timesteps.}
    \label{relative-timesteps}
\end{figure}

We extract the final layer of attention weights from the Decision Transformer models trained on the four reward scales for the three types of inputs: returns, states, and actions. We observe both the attention weights for the individual input types as well as the aggregated and averaged set of weights aross all input types. Due to the auto-regressive nature of Decision Transformer, attention weights are assigned to the \textit{timesteps} prior to the recommendation made at a given timestep.

We provide visual analyses of attention scores through normalized aggregate attention scores per timestep for absolute timestep values (Fig. \ref{absolute-timesteps}) as well as relative timestep values (Fig. \ref{relative-timesteps}). We note that the model refines its attention to generally focus on items in earlier positions in given input sequence, both in the case for absolute and relative timesteps. These results, in tandem with the generally higher performances of the model on 15 previous timesteps rather than 20 timesteps, indicate that potentially there is a beginning index to the current context that can be key for the model's inference ability. Future work may include adjusting the context window dynamically, both for training and inference.

\section{Limitations}

Due to limitations of computational resources,  experimentation with fine-tuning LLMs is restricted by model size. Future work can build on this work by applying similar experiments on increasing model sizes or non-quantized versions of models, effectively demonstrating (positively or negatively) that performance scales with model size.

\section{Ethical Considerations}

When implementing a topic recommendation system in counseling contexts, ethical considerations are important due to the sensitive nature of digital mental health discussions, as discussed in \cite{lin2022computational}. One of the primary concerns is the potential limitation imposed by a static set of discussion topics. While such a system can streamline the counseling process, it risks limiting the creativity and flexibility of counselors, particularly those in training, and in the long term, inhibit consideration of their own perspectives on how to continue the conversation. This could inadvertently restrict their ability to tailor sessions according to the unique needs of each patient.

This is particularly relevant since the topics pulled are from one specific dataset that covers only four mental health conditions. The training dataset, derived from this limited number of mental health conditions, might not be representative of the broader population or other conditions. This limitation can lead to biased recommendations if not carefully managed. To mitigate this, it is essential to consider a more dynamic approach where the set of topics can evolve based on ongoing input from practicing counselors and feedback from therapy sessions. This adaptation would help in maintaining the relevance and sensitivity of the recommendations to diverse patient needs. In deployment, we can also imagine that topics are dynamically chosen, or chosen using human feedback; for example, perhaps before the system is put into use, counselors can input their own topics.

In addition to dataset limitations, the calculation of rewards, based on the Working Alliance Inventory (WAI), while rooted in established psychological theory, may benefit from enhancements through reinforcement learning with human feedback (RLHF). Incorporating direct input from users could refine the understanding and alignment of counselor and patient goals, improving the system's effectiveness and ethical alignment.

\section{Conclusion}

In this study, we introduced a Decision-Transformer-based recommendation system which outperforms baseline RL-based methods in counseling topic recommendation, indicating that transformer-based methods may have better performance in general when it comes to modeling conversation direction and alignment. We additionally find that the model performs best for certain reward scales on shorter input sequences, indicating that some exploration of optimal sequence length can be an avenue for future work.
Through additional analysis of the attention scores, we additionally find that the model pays more attention to items earlier on in the input sequence.

\bibliography{anthology,custom}
\bibliographystyle{acl_natbib}

\appendix


\end{document}